%% file: main.tex
\documentclass[10pt, twocolumn]{article}
\usepackage[margin=0.75in]{geometry}
\usepackage{amsmath, amssymb, amsfonts}
\usepackage{graphicx}
\usepackage{booktabs}
\usepackage{abstract}

\usepackage{cite}
\usepackage{algorithm}
\usepackage{algorithmic}
\usepackage{textcomp}
\usepackage{enumitem}
\usepackage{orcidlink}

\usepackage{multirow}
\usepackage{siunitx}
\usepackage{array}
\usepackage[table]{xcolor}
\usepackage{colortbl}
\usepackage{pifont}
\usepackage{subcaption}
\usepackage{comment}

\definecolor{usedhighlight}{RGB}{255,105,180}
\definecolor{hotpink}{RGB}{252,214,243}

\providecommand{\IEEEmembership}[1]{\textnormal{\textit{#1}}}

\newcommand{\blfootnote}[1]{%
  \begingroup
  \renewcommand{\thefootnote}{}\footnote{#1}%
  \addtocounter{footnote}{-1}%
  \endgroup
}

\usepackage{hyperref}

\title{\textbf{Whole-Slide Image Analysis under Realistic Few-Shot Annotation Protocols}}

\author{
    Tiffanie Godelaine\,\orcidlink{0009-0006-2873-2686}, Maxime Zanella\,\orcidlink{0009-0009-4030-3704}, Karim El Khoury\,\orcidlink{0000-0002-1594-6531} \IEEEmembership{Member, IEEE},\\ Beno\^{i}t Macq\,\orcidlink{0000-0002-7243-4778} \IEEEmembership{Fellow, IEEE}, and Christophe De Vleeschouwer\,\orcidlink{0000-0001-5049-2929} \IEEEmembership{Member, IEEE}
}

\date{}

\begin{document}

\twocolumn[
\begin{@twocolumnfalse}
\maketitle

\begin{onecolabstract}
Automating the analysis of whole-slide images has high clinical value, since characterizing cancers requires examining them in detail.
Such analysis increasingly relies on vision-language models that provide patch-level zero-shot predictions.
However, these predictions remain noisy and must be refined with a few annotations.
A promising paradigm for this refinement is few-shot transduction. Rather than treating each patch independently, these methods leverage the relations between patches, together with a few annotations, to refine all predictions jointly. However, current transductive methods are evaluated under conditions that overlook key properties of whole-slide images: (i) datasets consist of independent patches extracted from multiple slides, ignoring the complex tissue organization; (ii) datasets are mostly balanced, whereas a single whole-slide image exhibits severe class imbalance, with several classes absent; and (iii) annotations are sampled at random, without reflecting how a pathologist annotates a limited number of regions.
To align the transduction paradigm to realistic whole-slide settings, we introduce the following contributions. First, we propose SlideCRF, which adapts conditional random fields for whole-slide images by combining spatial and biological cues while accounting for classes that may be absent from a given slide.
Second, we provide a set of realistic annotation protocols, based on spatially localized \emph{clicks} and \emph{scribbles}, modeling different pathologist interactions, such as the iterative correction of model errors.
Across four datasets, we show that SlideCRF outperforms current transductive methods in macro F1, improving over the zero-shot predictions by $+24.2\%$ and $+37.5\%$ with one and 16 clicks per present class, respectively.
\end{onecolabstract}

\vspace{2mm}
\vspace{6mm}
\end{@twocolumnfalse}
]

\blfootnote{Manuscript submitted on 29th July 2026.}
\blfootnote{Tiffanie Godelaine, Maxime Zanella, Karim El Khoury, Beno\^{i}t Macq and Christophe De Vleeschouwer are affiliated with the Institute of Information and Communication Technologies, Electronics, and Applied Mathematics (ICTEAM), Universit\'e catholique de Louvain (UCLouvain), 1348 Louvain-la-Neuve, Belgium.}
\blfootnote{Tiffanie Godelaine is supported by MedReSyst as part of the ``Wallonia 2021--2027'' program, Maxime Zanella is supported by ARIAC by DIGITALWALLONIA4.AI (grant No. 2010235) and Christophe De Vleeschouwer by the Fonds de la Recherche Scientifique (FNRS).}
\blfootnote{Computational resources have been provided both by the supercomputing facilities of the Universit\'e catholique de Louvain (CISM/UCL) and the Consortium des Equipements de Calcul Intensif en F\'ed\'eration Wallonie Bruxelles (CECI) under convention 2.5020.11 by the Walloon Region and the Fond de la Recherche Scientifique de Belgique (FNRS).}
\blfootnote{Corresponding e-mail: tiffanie.godelaine@uclouvain.be. Code and data are publicly available at: \url{github.com/tgodelaine/SlideCRF}.}

\section{Introduction}
\label{sec:introduction}
\input{Sections/0-introduction-new}

\section{Related Work}
\input{Sections/1-related_work}

\section{SlideCRF}
\input{Sections/2a-method}

\section{Annotation protocols}
\label{sec:annotation_protocols}
\input{Sections/2b-method}

\section{Experimental Setup}
\input{Sections/3-experimental_setup}

\section{Experiments}
\label{sec:exp}
\input{Sections/4-results}

\section{Conclusion}
\input{Sections/5-conclusion}

\bibliographystyle{IEEEtran}
\bibliography{biblio_3}

\end{document}

%% file: Sections/0-introduction-new.tex
Accurate analysis of whole-slide images (WSIs) is crucial for cancer diagnosis and staging, enabling pathologists to characterize complex histopathological entities. Automating this pipeline offers substantial clinical value by reducing workload while enhancing diagnostic capabilities~\cite{Moscalu2023}. However, given the gigapixel scale of WSIs, they must first be divided into independent, non-overlapping patches to make processing computationally feasible.

To analyze these massive patch collections, recent frameworks have increasingly relied on histology-specific vision-language models (VLMs)~\cite{conch, plip, quilt}. Pre-trained on massive histological image-text pairs, these models exhibit strong generalization capabilities that enable patch-level zero-shot (ZS) predictions without requiring prior annotations from pathologists. Yet, despite their promising performance, these initial ZS predictions often remain noisy, still necessitating a limited number of expert annotations for refinement~\cite{vlm_refinements}.

In histology, the most commonly used approaches to refine these VLM predictions are inductive few-shot methods~\cite{Lee2025}. These methods typically rely on few annotations to train a lightweight linear classification layer or optimize a small set of adapter parameters on top of a frozen VLM backbone, and generate an independent prediction for each patch~\cite{tip, coop, cliplora}. In contrast, a second and promising direction involves transductive few-shot methods~\cite{transductive_histo, histo-transclip}. Rather than treating each patch independently, these techniques jointly refine all predictions by leveraging both expert annotations and the underlying relationships between patches~\cite{thortsen1999, lp}. %, typically without retaining a model~\cite{transductive_histo, histo-transclip}. 
Conditional random fields (CRFs) fit into this paradigm: by modeling pairwise dependencies between patches, they enforce consistency across neighboring predictions while refining all patches at once~\cite{histocrf}.

\begin{figure*}[t]
    \centering
    \includegraphics[width=\linewidth]{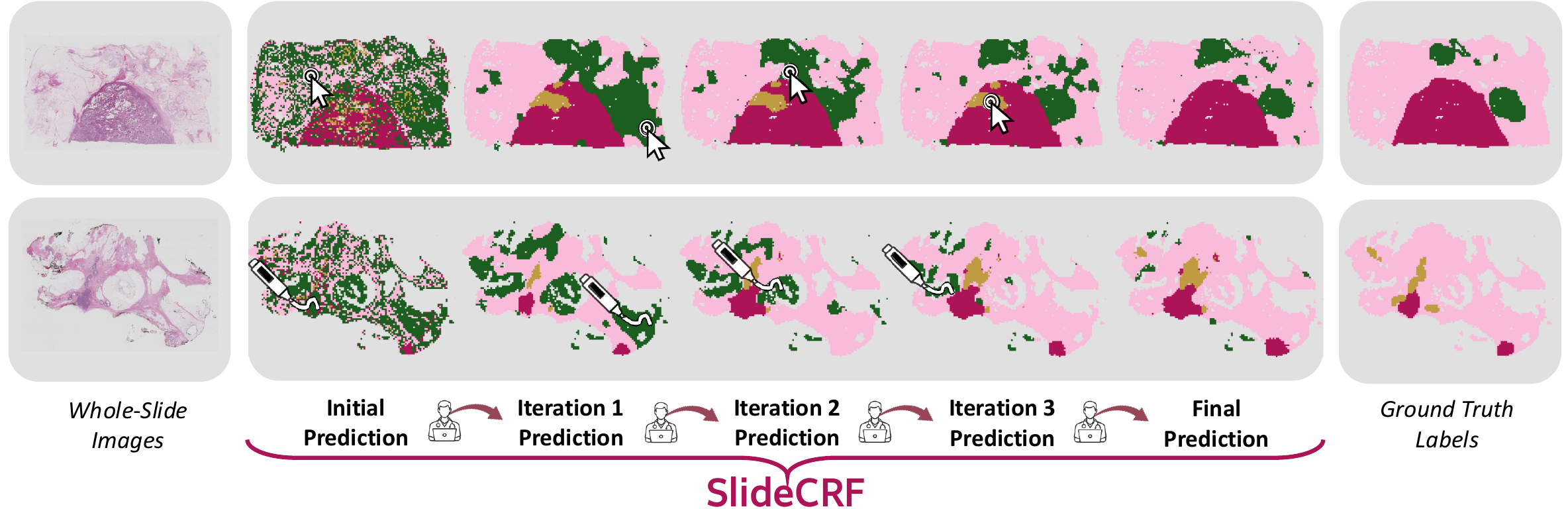}
    \caption{Illustration of SlideCRF on two WSIs. From a few localized \emph{clicks} (top) or a few localized \emph{scribbles} (bottom), SlideCRF progressively refines the prediction and recovers spatially coherent regions. \label{fig:illlustration}}
\end{figure*}

Despite their conceptual advantages in modeling patch dependencies, current state-of-the-art transductive methods, such as Histo-TransCLIP~\cite{histo-transclip} and HistoCRF~\cite{histocrf}, are evaluated under conditions that diverge sharply from realistic clinical settings. This discrepancy manifests in three main ways: (i) standard benchmarks use independent patches pooled from distinct WSIs, discarding the critical spatial context necessary to characterize complex tissue structures; (ii) these benchmarks rely on mostly balanced datasets, overlooking the severe intra-slide class imbalance and frequently absent diagnostic categories inherent to WSIs; and (iii) existing methods assume that expert annotations are sampled randomly across the entire slide. This ignores actual pathological workflows, where clinicians typically annotate a limited number of localized, contiguous regions rather than a sparse, widespread set of patches. Consequently, because existing transductive methods were not designed with these constraints in mind, their performance remains severely limited when evaluated under realistic clinical conditions.\\

To align transductive learning with realistic clinical settings, we introduce the following key contributions:
\begin{itemize}[leftmargin=*]
    \item We propose \textbf{SlideCRF}: a method to refine the patch-level ZS predictions of a WSI using a few expert annotations by modeling the relations between patches with a CRF. To handle the realistic WSI setting, our method integrates spatial and biological relations, recovering finely structured regions while remaining robust to the absence of classes within a slide. 
    As shown in Figure \ref{fig:illlustration}, by iteratively annotating a few regions, SlideCRF progressively refines the ZS predictions and recovers the underlying structures of the slide.%\\
    \vspace{1mm}
    \item We provide a set of \textbf{realistic annotation protocols} based on spatially localized \emph{clicks} and \emph{scribbles}, rather than widespread annotations, better reflecting how a pathologist interacts with a WSI. To capture realistic human-in-the-loop (HITL) dynamics, these annotations can target representative regions, focus on erroneous regions, or iteratively correct errors. Studying how each protocol affects refinement offers a rigorous baseline to benchmark future progress in interactive digital pathology. 

\end{itemize}

%% file: Sections/1-related_work.tex
\begin{comment}
\subsection{FS VLM adaptation}
\subsubsection{In natural images}
An approach to refine predictions when a few labeled patches are available, few-shot learning can be used to adapt VLMs, such as prompt learning \cite{coop} or feature adapters \cite{clipadapter, tip}. 

\subsubsection{In histology}
 
However, VLM predictions are often noisy at the patch-level and thus require refinements \cite{vlm_refinements}. When only a few labeled patches are available, VLMs can also be adapted with lightweight few-shot methods such as prompt learning \cite{coop} or feature adapters \cite{clipadapter, tip}. 
\end{comment}

\subsection{Transductive VLM prediction}

Unlike inductive few-shot methods~\cite{coop, tip, cliplora} that classify each patch independently, few-shot transductive methods refine all test predictions jointly, exploiting the relationships between unlabeled samples in addition to a few annotations.

\subsubsection{In natural images} 
EM-Dirichlet~\cite{dirichlet} was the first transductive method to explicitly leverage the textual modality of VLMs alongside the image embeddings: it models the samples 
with a Dirichlet distribution and uses the VLM predictions to optimize its parameters without retraining the model. In the same spirit, TransCLIP~\cite{transclip} models the data distribution as a Gaussian mixture and iteratively refines the pseudo-labels obtained from VLMs, both without labels and in a few-shot variant. More recently, StatA~\cite{stata} introduced a transductive approach robust to test conditions, where the set of test samples may not contain classes anticipated by the model. Other methods rely on graph modeling: building upon ZLAP~\cite{zlap}, which performs label propagation on a graph constructed from text prompts and unlabeled samples, ECALP~\cite{ecalp} extends this framework to the few-shot setting by incorporating annotated samples into the graph and introduces a context-aware feature re-weighting that calibrates similarities prior to propagation. 
%ECALP, however, constructs its propagation graphs exclusively in feature space.

\subsubsection{In histology}
Histo-TransCLIP~\cite{histo-transclip} applies TransCLIP to histology, jointly refining patch-wise predictions from the distribution of patches in the latent space. 
Building on this transductive setting, HistoCRF~\cite{histocrf} formulates patch classification as a CRF inference problem and combines VLM predictions with a pairwise potential derived from a small set of annotations to refine all ZS predictions at once.
In contrast, PADDLE~\cite{transductive_histo} operates on the patches within a local window of the WSI and, from a few labeled patches, estimates the optimal class assignments of these patches while penalizing a large number of distinct predicted classes. This constraint reflects a key characteristic of the WSI setting: only a few of the many candidate classes are present in a given slide. \\ 
%https://iopscience.iop.org/article/10.1088/2632-2153/ae0557/meta

\textbf{Current limitations.} Despite their differences, except PADDLE which leverages spatial proximity by restricting its joint analysis on local windows, all the mentioned methods construct relationships primarily in the feature space, connecting patches according to their feature similarity, without using the raw data. In histopathology, tissue structures share biological information and exhibit strong spatial continuity. Therefore, relying solely on visual similarity may connect anatomically distant regions while overlooking local tissue organization. This motivates the integration of spatial and biological information into the refinement process, which we incorporate through a CRF combining spatial and biological cues in addition to model features.

\subsection{Annotation protocols}
A realistic WSI refinement method naturally raises the question of how these annotations are generated. 

\subsubsection{In medical images}
To reduce the burden associated with pixel-wise segmentation mask delineation, weak annotations have been widely adopted, including points~\cite{points_ann_histo_2}, scribbles~\cite{Wang2026} and bounding boxes~\cite{gaillochet2025prompt}, which trade annotation density for a much lower labeling cost.

To train methods that operate without dense supervision, several tools have been introduced to automatically generate sparse annotations from ground-truth masks~\cite{marinov2024survey}. For scribbles, ScribbleBench~\cite{scribblebench} samples control points inside a region and along its boundary, then interpolates a smooth curve between them. For clicks, points are sampled randomly within the object while satisfying minimum-distance constraints from both the boundary and previously sampled points, ensuring that they remain well distributed~\cite{xu2016deepinteractive}. Alternatively, clicks are placed on the erroneous regions of the current prediction, typically on the largest erroneous region~\cite{mahadevan2018itis, sofiiuk2022ritm, liu2023simpleclick}, optionally at the point furthest from its boundary~\cite{sakinis2019deepgrow}.
Applied iteratively, this correction strategy allows the prediction to be refined over successive rounds~\cite{sakinis2019deepgrow}. 

\subsubsection{In histology}
\cite{scribble_ann_histo} propose a method to generate scribbles from masks of tumoral regions. To simulate a pathologist drawing a continuous curve along a tumor, each generated scribble spans the whole region. 
Iterative correction has also been studied: \cite{lutnick2019iterative} propose an interface for data annotation that iteratively retrains a model from clicks provided by a pathologist, and \cite{scribble_ann_histo} study the interactive correction of erroneous regions, reaching high performance with only a few scribbles. More recently, interactive foundation models such as Vista-path~\cite{vistapath} propagate sparse expert feedback into whole-slide segmentation, but rely on a trained model. \\

\textbf{Current limitations.} Each of these tools captures a different facet of the annotation process but none combines them into a realistic benchmark covering a range of annotation budgets, types, and interactions.
To fill this gap, we propose realistic WSI annotation protocols that span two annotation types (clicks and scribbles) and three pathologist interactions such as the correction of errors. These protocols allow the generation of a range of annotation budgets and reflect a pathologist annotating distinctive regions.
This provides a unified benchmark of pathologist interactions, from which we derive practical recommendations for annotating WSIs. The annotation protocols are further detailed in Section~\ref{sec:annotation_protocols}.

%% file: Sections/2a-method.tex
\subsection{CRF Objective function}
We propose an adaptation of CRFs for realistic WSI analysis. To handle large-scale histopathological images, we divide them into a set of patches. The WSI is represented as a graph $\mathcal{G}=(\mathcal{V}, \mathcal{E})$, in which each patch corresponds to a vertex $v \in \mathcal{V}$ and relations between patches are represented by the set of undirected edges connecting the vertices $\mathcal{E}$ \cite{BenAyed2023}.

We want to obtain an assignment $\mathbf{y}=(y_v)_{v \in \mathcal{V}}$, defining a class label for each patch. In CRF, the joint distribution over the hidden random variables is commonly expressed as a Gibbs distribution:
\begin{equation}
    P(\mathbf{Y}=\mathbf{y}) = \frac{1}{Z}\exp(-E(\mathbf{y})), 
\end{equation}
where $Z$ is a normalizing factor and $E(\mathbf{y})=\sum_{c \in \mathcal{C}} \Phi_c(y_c)$ is the energy defined on the set of maximal cliques $\mathcal{C}$ (a clique is a fully connected sub-graph).

If we state our problem as a Bayesian estimation $P(\mathbf{Y}=\mathbf{y}|\mathbf{X}) \propto P(\mathbf{X}|\mathbf{Y}=\mathbf{y})P(\mathbf{Y}=\mathbf{y})$ and taking the negative logarithm, we obtain an objective function of the general form:

\begin{equation}
\label{eq:obj_fun}
\begin{split}
    f(\mathbf{y}) &= - \sum_{v\in\mathcal{V}} \log P(x_v|Y_v=y_v) + E(\mathbf{y})  \\
    &= \underbrace{\sum_{v \in \mathcal{V}} \phi_v (y_v)}_{\substack{\text{unary potentials $\Phi_u$}}} + \underbrace{\sum_{c \in \mathcal{C}} \Phi_c(y_c)}_{\substack{\text{pairwise potentials $\Phi_p$}}}.
\end{split}
\end{equation}
Unary and pairwise potentials are the building blocks of CRF methods. The former indicate the value that each variable is likely to take based on its own observed variables and the latter encourage neighboring variables to take compatible values.

\subsection{Unary potential}
In this work, we build on the definition of the unary potential proposed in our preliminary work \cite{histocrf}, and add a term to be robust to the possible absence of classes within a slide.

\subsubsection{Initial prediction term}
For this term, we rely on histology-oriented VLMs to obtain our unary potential $\phi_v(y_v)$ from patch-level predictions without training a task-specific model. More precisely, we encode each patch $v$ with the visual encoder to get an embedding vector $\mathbf{f}_v$ and for each class $l$ textual embeddings $\mathbf{t}_l$ with the textual encoder from the average of a set of prompts (e.g., \textit{``a histology image of \{class\}''}) \cite{histo-transclip}.
The unary potential is then computed from the cosine similarity:
\begin{equation}
\label{eq:unary}
\phi_v(y_v = l)
= - \log \left(
\operatorname{softmax}_l \left(
\frac{\mathbf{f}_v \mathbf{t}_l^\top}{\|\mathbf{f}_v\| \, \|\mathbf{t}_l\|}
\right)
\right).
\end{equation}

\subsubsection{Class-presence term}
The expert annotations $\mathcal{A}$ give an indication of which classes are actually present in the slide. We exploit this to discourage the prediction of classes that never appear in the annotation set. The prediction of a class is driven by its unary potential: a class with a low unary potential tends to be predicted. Let $\mathcal{P}$ be the set of annotated classes. To prevent the prediction of absent classes, we add a constant $\lambda$ to the unary potential of every unannotated class $l \notin \mathcal{P}$, at every patch $v$: 

\begin{equation}
\label{eq:unary_prior}
\tilde{\phi}_v(y_v = l)
= \phi_v(y_v = l) + \lambda \, \mathbb{1}\!\left[\, l \notin \mathcal{P} \,\right],
\end{equation}
where $\mathbb{1}[\cdot]$ is the indicator function.

\subsection{Pairwise potentials}
The definition of the pairwise potentials differs from that in our initial work \cite{histocrf} by the addition of two additional potentials to the annotation and diversity terms originally defined. 

\subsubsection{Annotation term} 
The first term connects each annotated patch $v \in \mathcal{A}$ to its set $\mathcal{M}_v$ of highly similar patches according to their embedding vectors $\mathbf{f}_v$, extracted using a histology-oriented vision model. This encourages the refinement process to align their labels with expert input, promoting consistency with pathologist annotations. 
This leads to the following definition of the first pairwise term: 
\begin{equation}
\label{eq:pairwise_ann}
\psi_{vw}(y_v,y_w)=(1-\delta_{y_v,y_w})\text{sim}(\mathbf{f}_v, \mathbf{f}_w),
\end{equation}

where similarities are measured using the cosine similarity: 
\begin{equation}
    \text{sim}(\mathbf{f}_v, \mathbf{f}_w) = \frac{\mathbf{f}_v \mathbf{f}_w^\top}{\|\mathbf{f}_v\|\|\mathbf{f}_w\|}. 
    \label{eq:simi}
\end{equation}

\subsubsection{Diversity term} 
For the second term, we connect each patch $v$ to its set $\mathcal{N}_v$ of most dissimilar patches. By linking patches that appear different but share the same label, this term encourages the refinement process to adjust such inconsistencies, promoting greater label diversity. 
\begin{equation}
\label{eq:pairwise_base}
\phi_{vw}(y_v,y_w)=\delta_{y_v,y_w}(1-\text{sim}(\mathbf{f}_v, \mathbf{f}_w)),
\end{equation}

\subsubsection{Spatial term} 
For the third term, we use spatial connections between patches by connecting each patch $v$ to the set $\mathcal{S}_v$ of its spatial neighbors. This term acts as a smoothing incentive that favors the assignment of identical labels to neighboring patches.
It takes the same functional form as the annotation term (Eq. \ref{eq:pairwise_ann}), but operates on a different set of pairs.

\subsubsection{Biological term}
The last pairwise term introduces handcrafted biological cues and encourages patches that appear different to the vision model but are biologically similar to receive the same label.
It combines two feature groups, texture $\mathbf{f}_1$ and color $\mathbf{f}_2$:
\begin{equation}
    \label{eq:pairwise_biological}
    \xi_{vw}(y_v, y_w)=(1-\delta_{y_v,y_w})\sum_{k=1}^{2}\eta_k\,
     \text{sim}_2(\mathbf{f}_{k,v}, \mathbf{f}_{k,w}),
\end{equation}
where the similarity is computed using a Gaussian kernel applied to the Euclidean distance between the features extracted from each patch: 
\begin{equation}
    \label{eq:sim2}
    \text{sim}_2(\mathbf{f}_{k, v}, \mathbf{f}_{k, w}) = \text{exp}
    \left(
    -\frac{||\mathbf{f}_{k, v} - \mathbf{f}_{k, w}||^2}{2\sigma^2}
    \right). 
\end{equation}

Texture concatenates gray-level co-occurrence matrix~\cite{haralick1973} (contrast, homogeneity, energy, correlation) and local binary patterns~\cite{ojala2002}, capturing the spatial arrangement of pixels. Color features~\cite{ruifrok2001} are computed on image patches converted in the hematoxylin-eosin-DAB (HED) color space, and concatenate the mean and standard deviation of the channels corresponding to the hematoxylin and eosin stains.

\textbf{Combined pairwise potential.} Combining the four terms gives our pairwise potential:\\
\begin{multline}
\label{eq:pairwise_potential}
\Phi_p =
    \underbrace{\beta \sum_{v \in \mathcal{A}} \sum_{w \in \mathcal{M}_v} \psi_{vw}(y_v,y_w)}_{\text{(i) annotation pairwise term}}
+
    \underbrace{ \alpha\sum_{v \in \mathcal{V}} \sum_{w \in \mathcal{N}_v} \phi_{vw}(y_v, y_w)}_{\text{(ii) diversity pairwise term}}
\\ + 
    \underbrace{\gamma \sum_{v \in \mathcal{V}} \sum_{w \in \mathcal{S}_v} \psi_{vw}(y_v,y_w)}_{\text{(iii) spatial pairwise term}}
+ 
    \underbrace{\sum_{v \in \mathcal{V}} \sum_{w \in \mathcal{N}_v} \xi_{vw}(y_v,y_w)}_{\text{(iv) biological pairwise term}}
,
\end{multline}
where $\alpha$, $\beta$, $\gamma$ and $\eta$ are weighting factors. 
To be robust to the partial presence of certain classes, $\alpha$ follows a step schedule: $\alpha_i = \alpha$ if $i < \tau$ and $\alpha_i = 0$ otherwise, where $i$ is the message passing iteration (see next section \ref{sec:opti}) and $\tau$ is a fixed threshold.
The sets $\mathcal{N}_v$ and $\mathcal{M}_v$ are not fixed but are drawn at random from within a subset of the most dissimilar or most similar patches to $v$ each time the pairwise potential is computed to increase the number of connections without increasing memory load.

\subsection{Objective function optimization}
\label{sec:opti}
Combining the definitions of both potentials gives the following objective function: 
\begin{align}
    \label{eq:obj}
    f(\mathbf{y}) = \sum_{v \in \mathcal{V}} \tilde{\phi}_v(y_v) + \Phi_p
\end{align}

As the solution of Eq. \eqref{eq:obj} is intractable, a mean-field approximation is used. This minimizes the KL-divergence between %the CRF distribution 
$P(\mathbf{Y}|\mathbf{X})$ and a simpler approximation $Q$ that can be written as the product of independent marginal distributions  $Q(\mathbf{y)}=\prod_v Q_v(y_v)$. 
This results in an iterative update equation corresponding to a message passing algorithm \cite{Krahenbuhl2012} on the graph  that can be written as:
\begin{equation}
\label{eq:up}
\begin{split}
Q_v(l) = \frac{1}{Z_v}\exp\Big(
&-\tilde{\phi}_v(l)
- \alpha\!\!\sum_{w\in\mathcal{N}_v}\!\!\big(1-s_{vw}\big)Q_w(l)\\[-1pt]
&- \!\!\sum_{\substack{l'\in\mathcal{L}\\ l'\neq l}}\!\Big[
\beta\!\!\sum_{w\in\mathcal{M}_v}\!\! s_{vw}\,Q_w(l')
+ \gamma\!\!\sum_{w\in\mathcal{S}_v}\!\! s_{vw}\,Q_w(l')\\[-1pt]
&+ \!\!\sum_{w\in\mathcal{N}_v}\!\! s_{2,vw}\,Q_w(l')
\Big]\Big),
\end{split}
\end{equation}
where $s_{vw}=\mathrm{sim}(\mathbf{f}_v,\mathbf{f}_w)$,
$s_{2,vw}=\sum_{k}\eta_k\,\mathrm{sim}_2(\mathbf{f}_{k,v},\mathbf{f}_{k,w})$,
and $Q_w(l)\equiv Q_w(y_w=l)$.

The key steps of the method are outlined in Algorithm~\ref{alg:it}.

\input{Tables/algorithm}

%% file: Tables/algorithm.tex
\begin{algorithm}[t]
\resizebox{0.9\linewidth}{!}{%
\begin{minipage}{\linewidth}
\caption{SlideCRF (HITL)}
\label{alg:it}

\textbf{For} $1 \to \text{Number of expert-annotation steps}$

\hspace*{1em} Get pathologist annotations $\mathcal{A}$

\hspace*{1em} Initialize unary potential $\tilde{\phi}^{(0)} \leftarrow \text{Eq.}~\eqref{eq:unary_prior}$

\colorbox{hotpink}{%
\parbox{\dimexpr\linewidth-2\fboxsep}{%
    \textcolor{black}{\hspace{0.3cm}\textit{Iterative message passing}}\\
    \hspace*{1em} \textbf{For} $i = 1 \to 50$\\ %\footnote{This \textit{for} loop corresponds to one iterative message passing.}\\
    \hspace*{2em} Get pairwise potential $\Phi_p \leftarrow \text{Eq.}~\eqref{eq:pairwise_potential}$\\
    \hspace*{2em} Iterative update: $Q^{(i)} \leftarrow \text{Eq.}~\eqref{eq:up}$\\
    \hspace*{2em} Update unary potential: $\phi^{(i)} \leftarrow \frac{1}{2}\left(\phi^{(i-1)} + Q^{(i)}\right)$
}}
\textbf{Output:} Predictions $\hat{y} \leftarrow \operatorname*{arg\,max} Q^{(50)}$ \\
\end{minipage}
}
\vspace{-2ex}
\end{algorithm}

%% file: Sections/2b-method.tex
%Since collecting real pathologist annotations at multiple budgets is time-consuming, we simulate expert annotations from the ground-truth segmentation masks. We consider two annotation modalities commonly used in medical imaging: clicks~\cite{points_annotation} and scribbles~\cite{scribblebench} (Fig.~\ref{fig:visu_annotations}). 
%We focus on two weak-annotation modalities, \emph{clicks} and \emph{scribbles}, as they are the most widely used for region-level labeling in histopathology and offer a favorable accuracy/effort trade-off: dense masks are prohibitively costly at whole-slide scale, while coarser cues such as bounding boxes are ill-suited to patch-level supervision because their axis-aligned extent spans several tissue types.

\begin{figure}[!b]
    \centering
    \includegraphics[width=\linewidth]{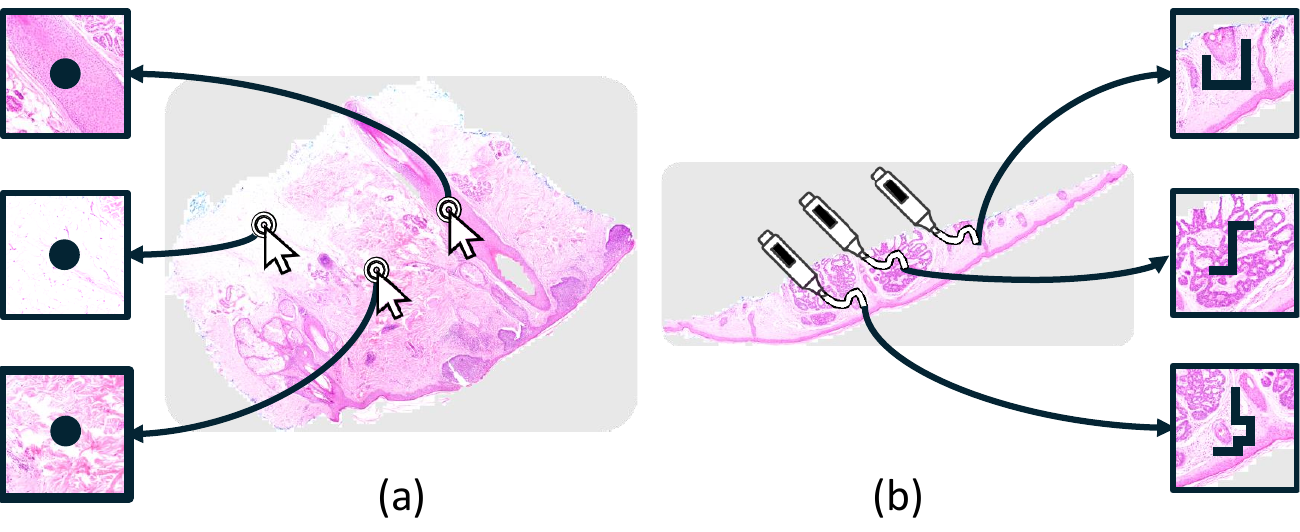}
    \caption{Examples of both annotation types, (a) \textit{clicks} and (b) scribbles, on two WSIs.}
    \label{fig:visu_annotations}
\end{figure}

An annotation protocol is defined by two components: an annotation type, which determines the shape of the annotation, and an interaction type, which determines how the pathologist selects the patches to annotate. 

\subsection{Annotation types}
We consider two annotation types commonly used in medical imaging: \emph{clicks} and \emph{scribbles} (mainly based on ScribbleBench~\cite{scribblebench}).
An example of both
annotation types is shown in Figure 2.

\subsubsection{Clicks}
A \emph{click} is defined by a single control point (the center) and covers a disk of radius one patch around it, yielding five annotated patches (the center and its four neighbors). 

\subsubsection{Scribbles}
A \emph{scribble} is defined by several control points. A smooth curve, whose length is capped at 15 patches, is then interpolated between these points.\\ %The control points are sampled within the connected component of the target class, drawn preferentially towards its interior, so that the scribble stays inside a homogeneous tissue region.

\textbf{Generation process.} In our experiments, the center of a \emph{click} and the control points of a \emph{scribble} are generated automatically from the dense segmentation mask by a three-step process, designed to mimic a pathologist annotating a region of a tissue class:

\begin{enumerate}[leftmargin=*]
    \item[\textit{(i)}] \textit{Component selection.} For each class present in the slide, its connected components are extracted and one is sampled with probability proportional to its area, reflecting the tendency to annotate the largest, most visible structure. 
    \vspace{1mm}

    \item[\textit{(ii)}] \textit{Control point sampling.} Within the selected component, a position is sampled from a Gaussian distribution. Its center is placed at the point furthest from the boundary (identified by the Euclidean distance transform), and its standard deviation is set to $0.4$ times that point's distance to the boundary. %to prevent the sampling of the position at the boundary of the component
    The Gaussian is further modulated by the model confidence, reflecting the tendency to annotate central and representative regions. For \emph{scribbles}, the remaining control points are determined following ScribbleBench. 
    
 \vspace{1mm}

    \item[\textit{(iii)}] \textit{Repulsion.} When several annotations fall in the same region, the sampling probability is weighted by the distance from already-annotated patches. This favors successive annotations to cover different parts of the structure.
\end{enumerate}

\subsection{Interaction types}
We consider three pathologist interactions: \emph{representative}, \emph{error-driven} and human-in-the-loop (\emph{HITL}).

\subsubsection{Representative}
Annotations target the most representative region of each class, following the three-step generation process introduced in the previous subsection. 

\subsubsection{Error-driven}
Annotations correct erroneous patches predicted by the model. The control points (step \textit{(ii)} of the generation process) are drawn exclusively among the mispredicted patches of the selected component. Each annotation is thus anchored on an error. 

\subsubsection{Human-in-the-loop (HITL)}
Rather than spending the entire annotation budget at once, as in the \emph{representative} and \emph{error-driven} interactions, one annotation per class is added after each inference round.
As in the \emph{error-driven} interaction, each new annotation is anchored on a currently mispredicted patch, so that successive annotations progressively target the remaining errors of the model. 

%% file: Sections/3-experimental_setup.tex
\input{Tables/datasets}

\subsection{Datasets}
We evaluate the proposed method on four diverse histology WSI segmentation datasets: BACH~\cite{bach}, CATCH~\cite{catch}, SKINCANCER~\cite{skincancer} and TIGER~\cite{tiger2022challenge, amgad2019structured}. 
These datasets span a wide range of cancer types and species, 
and they differ substantially in the number of classes, slide sizes and spatial complexity. 
This diversity allows us to assess the robustness of the proposed method across heterogeneous segmentation settings.
For all datasets, dense segmentation masks are available, allowing patch-based evaluation.

\subsection{Patch extraction}
Each slide is tiled into non-overlapping patches ($448{\times}448$ px, reduced to $64{\times}64$ for SKINCANCER, whose slides are sometimes considerably smaller). Background tiles are discarded with a per-dataset tissue filter: a grayscale standard deviation threshold ($<0.025$) for BACH, a low-saturation HSV criterion for CATCH, and a tissue-mask coverage ratio ($>0.5$) for SKINCANCER and TIGER. Each retained patch receives a single label. For datasets annotated with polygons (BACH, TIGER) it is the first polygon intersecting the patch, defaulting to \emph{healthy} when none applies. For the densely-masked datasets (CATCH, SKINCANCER) it is the pixel-majority class within the patch. This yields between $\sim 950$ to $\sim 35,000$ labeled patches per slide, reflecting the wide range of slide sizes across datasets. A summary of the datasets' characteristics is presented in Table \ref{tab:data_characteristics}.

\subsection{Implementation details}
Each slide is segmented independently: the model has access to all of its unlabeled patches and to a set of annotated patches sampled according to the protocols of Section~\ref{sec:annotation_protocols}.
We vary the annotation budget $b \in \{1,2,4,8,16\}$. 
It is expressed in annotation \emph{actions} so that protocols compared at equal budgets involve the same number of annotation actions.
A single click and a single scribble each count as one action, and a budget of $b$ corresponds to $b$ actions per class present in the slide.
As in \cite{histocrf}, we use CONCH~\cite{conch} to define the unary potential, while UNI-2h~\cite{uni} provides the features used to build the pairwise potential.  
The weighting factors are set to $\alpha = 0.1$, $\beta = \gamma = 0.5$, $\eta_1 = 0.2$ (texture), $\eta_2 = 0.1$ (color), and the Gaussian kernel defining the biological affinity uses a standard deviation of $\sigma = 0.1$. The pairwise potential terms are sparsified to $|\mathcal{N}_v| = 4$ and $|\mathcal{M}_v| = |\mathcal{S}_v| = 8$ connections per patch.
Mean-field inference is run for 50 propagation steps; the diversity term is removed at $\tau=25$ (i.e., halfway through inference) and the class-presence prior is set to $\lambda=0.02$. These hyperparameters were determined on a validation set of 20 slides randomly sampled from SKINCANCER (the largest dataset) and excluded from the test set. These hyperparameters are fixed once and kept identical across all runs on all four datasets, which is consistent with our realistic setting. An ablation study (Section~\ref{sec:ablation}) evaluates the contribution of each term in the pairwise potential.

\subsection{Baselines}
We compare SlideCRF against two few-shot \emph{inductive} baselines, linear probe and Tip-Adapter-F, and four \emph{transductive} ones. Among the transductive methods, we distinguish two families. \emph{Probabilistic} approaches (Histo-TransCLIP, PADDLE) fit a distribution over the unlabeled patches and refine all predictions jointly through this global distribution. \emph{Graph-based} approaches (ECALP, HistoCRF) instead build a graph over patches and propagate labels along its edges, so that refinement is driven by explicit pairwise relations. 
Among all baselines, ECALP is the closest to our setting, as it also combines a few annotations with VLM textual embeddings for label propagation; SlideCRF differs by additionally modeling spatial and biological pairwise potentials within a CRF.
We additionally report the ZS CONCH prediction as a reference baseline. Whenever available, official implementations and recommended hyperparameters are used for each baseline.

\subsection{Evaluation metrics}
Performance is measured with the balanced accuracy and the macro F1. 
Balanced accuracy is the average, over the classes present in the slide, of the per-class recall, i.e., the fraction of each present class's patches that are correctly retrieved.
As it averages over the classes present in the slide, balanced accuracy does not directly take into account the prediction of classes that are absent from the slide. 
Macro F1 is the average F1 score of each class, which combines recall with precision, the fraction of patches predicted as a class that truly belong to it. 
Both metrics are computed per slide and then averaged over all evaluated patients in the dataset. We repeat each experiment over five random seeds, with each seed drawing a different set of annotations. 

%% file: Tables/datasets.tex
\begin{table}[t]
    \centering
    \footnotesize

    \resizebox{\linewidth}{!}{%
    \begin{tabular}{lccccc}
        \toprule
        \textbf{Dataset} & \textbf{Target organ} & \textbf{\#WSI} & \textbf{\#Cls.} & \textbf{Patch size} & \textbf{Patches/slide} \\
        \midrule
        BACH       & Breast (human)   & 10  & 4  & $448^2$ & $1.2$–$4.7$k \\
        CATCH      & Skin (canine)    & 70  & 12 & $448^2$ & $4.2$–$17.8$k \\
        SKINCANCER & Skin (human)     & 290 & 11 & $64^2$  & $0.9$–$35.0$k \\
        TIGER      & Breast (human)   & 93  & 2  & $448^2$ & $1.9$–$10.0$k \\
        \bottomrule
    \end{tabular}%
    }
    \caption{Dataset characteristics. \#Cls. is the number of tissue
    classes and Patches/slide is the min–max number of extracted
    patches per slide. Patch sizes are in pixels.}
    \label{tab:data_characteristics}
\end{table}

%% file: Sections/4-results.tex
\subsection{SlideCRF matches or outperforms baselines}

\input{Tables/table_bacc_f1}

Table~\ref{tab:bacc_f1} summarizes the comparison against the baselines. 

SlideCRF delivers substantial gains over the ZS prediction: with only one annotation per class present in the slide, it already improves by $+29.0\%$ in balanced accuracy and $+24.2\%$ in macro F1, and these gains keep growing with the annotation budget (up to $+43.5\%$ and $+37.5\%$, respectively).

\vspace{1.5mm}

The two top-performing approaches are SlideCRF and ECALP. Both of these methods leverage transductive graph-based learning and deliver highly competitive results. While SlideCRF secures the highest average score across both metrics, driven by a more pronounced lead in macro F1, it performs nearly on par with ECALP in balanced accuracy.

\vspace{1.5mm}

Beyond predictive performance, these methods differ significantly in computational cost (Table~\ref{tab:time_transposed}). SlideCRF processes a slide of approximately $10^5$ patches in about 38 seconds, remaining 13 times faster than ECALP. 

\vspace{1.5mm}

Within the graph-based methods, SlideCRF improves substantially over HistoCRF~\cite{histocrf}, showing that adding spatial and biological information, together with the class-presence prior, raises performance by $+10.3\%$ in balanced accuracy and $+12.4\%$ in macro F1 at an annotation budget of one.

\vspace{1.5mm}

Both transductive probabilistic methods exploit annotations poorly. Histo-TransCLIP shows almost no sensitivity to the number of annotation actions, suggesting that it makes little use of the expert-provided annotations (its average balanced accuracy increases by only $+1.2\%$ between $b=1$ and $b=16$). PADDLE does benefit from additional annotations but remains far below the graph-based methods on every dataset.

\vspace{1.5mm}

Among inductive methods, Linear Probe gains little at low budgets. In contrast, Tip-Adapter-F leverages the annotations effectively: its performance increases substantially as the number of annotated patches grows, making it the strongest non-graph baseline at large budgets.
\vspace{1.5mm}

\begin{figure}[t]
    \centering
    \includegraphics[width=\linewidth]{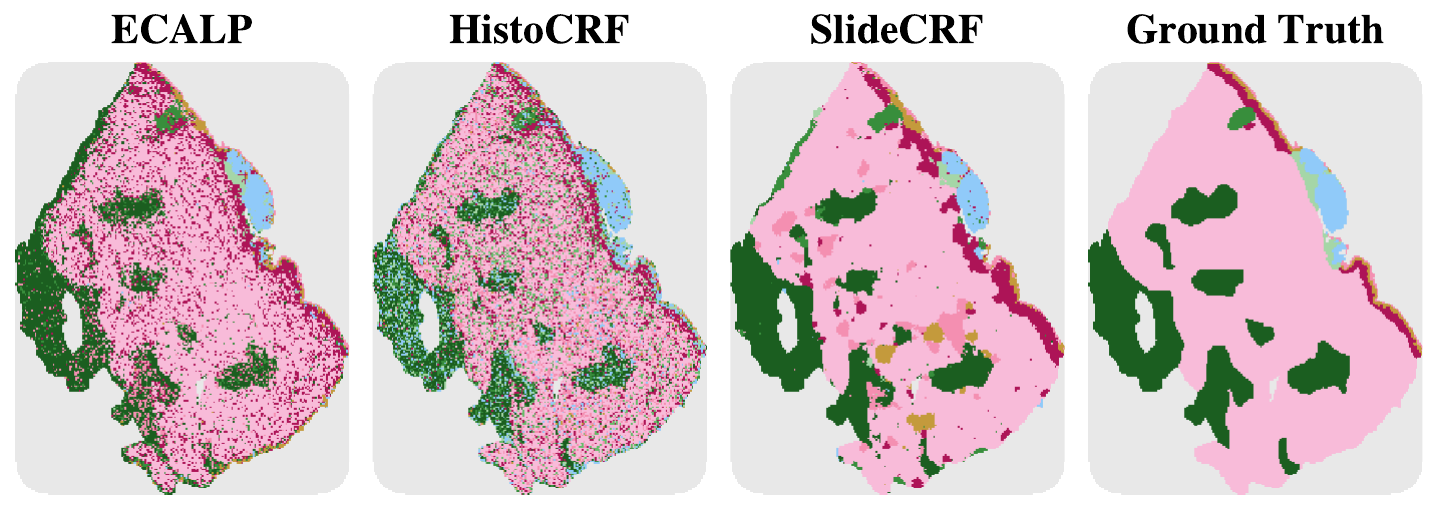}
    \caption{Qualitative comparison of graph-based methods on a SKINCANCER slide with $b=4$ \emph{clicks} (in the \textit{representative} interaction). 
    SlideCRF recovers spatially coherent regions, whereas the other methods ECALP and HistoCRF leave noisier, fragmented boundaries.}
    \label{fig:visualization_baselines}
\end{figure}

Visually (see Figure \ref{fig:visualization_baselines}), ECALP and HistoCRF produce noisier predictions at region borders, as they rely solely on embedding similarities and ignore explicit spatial information, whereas SlideCRF enforces spatial smoothness through its spatial pairwise potential, yielding cleaner and more coherent regions. 

\input{Tables/runtime}

\subsection{Ablation studies}\label{sec:ablation}

We evaluate the contribution of the novel components of our method: the spatial and biological terms, % added to the pairwise potentials, 
the weighting factor $\alpha$, % of the diversity term, 
and the class-presence term $\lambda$. Unless stated otherwise, all ablations use an annotation budget $b=4$ \emph{clicks} (in the \textit{representative} interaction) and $\lambda=0.02$.

\subsubsection{Spatial and biological terms}

Table~\ref{tab:ablation_potentials} highlights the improvements in balanced accuracy and macro F1 achieved by incorporating spatial and biological terms into SlideCRF, compared to the HistoCRF baseline.

\vspace{2mm}

\begin{figure}[!h]
    \centering
    \includegraphics[width=0.97\linewidth]{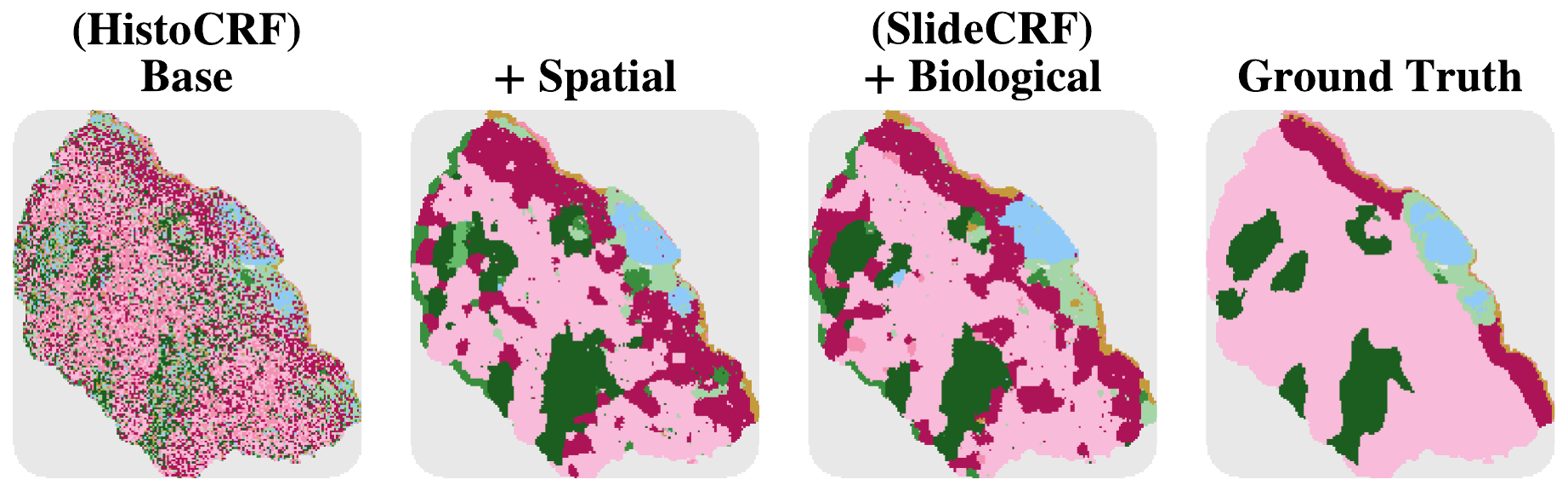}
    \caption{Effect of the addition of the spatial and biological pairwise terms on a SKINCANCER slide. % with $b=4$ \emph{clicks} (in the \textit{representative} interaction). 
    The spatial term recovers more coherent tissue regions, while the biological term sharpens class boundaries.}
    \label{fig:ablation_pair_pot}
\end{figure}

The spatial term is by far the largest contributor, yielding gains of $+5.9\%$ in balanced accuracy and $+8.1\%$ in macro F1. 
This is expected, since tissue in a WSI is spatially organized into coherent regions, making a patch's class highly predictive of its neighbors.
%neighboring patches tend to exhibit high embedding similarity.
As shown in Figure~\ref{fig:ablation_pair_pot}, the spatial term acts as a smoothing mechanism that promotes label consistency across neighboring patches.

\begin{figure}[!h]
    \centering
    \includegraphics[width=0.945\linewidth]{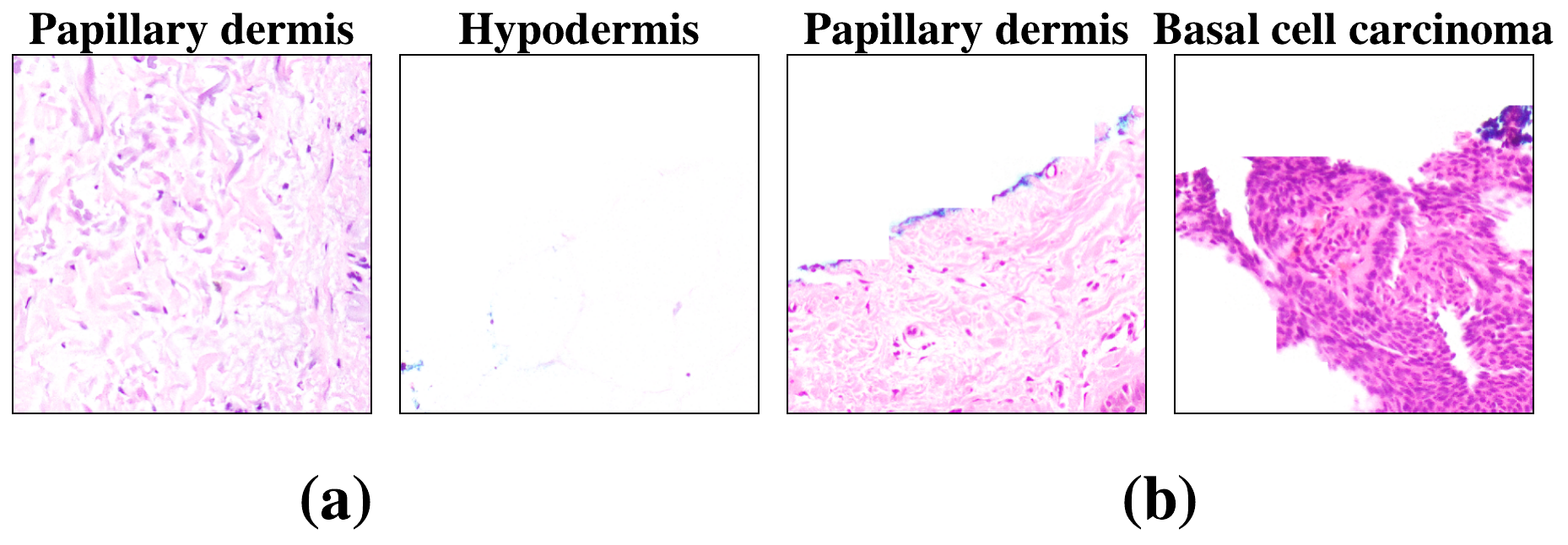}
    
    \caption{Visual difference on SKINCANCER patches (a) between papillary dermis (fine collagen) and hypodermis (empty, honeycomb-like fat vacuoles), separated by texture; 
    (b) between papillary dermis (pink collagen) and basal cell carcinoma (blue basaloid cells), separated by color.}
    \label{fig:ablation_bio_pair_pot}
\end{figure}

The biological term, combining texture and color cues, further improves macro F1 by $+5.4\%$. They allow us to refine the boundaries between the present classes. In  Figure~\ref{fig:ablation_pair_pot}, we can see that the biological term, in particular the texture, helps distinguish papillary dermis (light pink) from hypodermis (dark pink) classes. Although both correspond to pink tissue with few visible nuclei from a semantic perspective, their textures differ markedly as shown in the examples of Figure~\ref{fig:ablation_bio_pair_pot}. 
The color, in turn, helps distinguish papillary dermis from basal cell carcinoma: while papillary dermis is composed of pink collagen, basal cell carcinoma consists of blue basaloid cells. 
 
\subsubsection{Diversity weighting factor and class-presence term}
Table~\ref{tab:ablation_alpha} reports the effect of the weighting factor $\alpha$ of the diversity term and the class-presence term $\lambda$ on the balanced accuracy and the macro F1 score. 
The role of the diversity term is to promote label diversity, and therefore increase the number of predicted classes. Setting $\alpha$ uniform allows the propagation of classes missed by the ZS predictions. This explains the gains in both balanced accuracy and in macro F1, compared to the version without this term.

\vspace{2mm}

\begin{figure}[!t]
    \centering
    \includegraphics[width=0.97\linewidth]{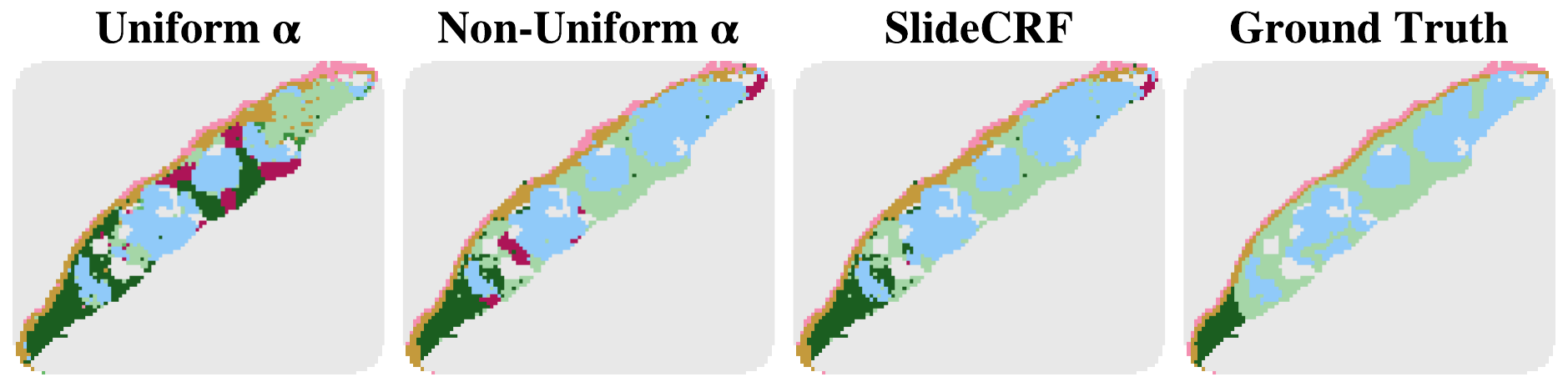}
    \caption{Effect of the mechanisms handling absent classes on a SKINCANCER slide. Setting $\alpha$ to 0 halfway to the propagation (non-uniform $\alpha$) and adding the class-presence term (SlideCRF) progressively suppress the prediction of absent classes, unlike a uniform $\alpha$.}
    \label{fig:ablation_alpha}
\end{figure}

However, in WSIs, some classes may be absent. We therefore adopt a step schedule that sets $\alpha$ to $0$ halfway through the propagation ($\alpha$ non-uniform), so as to mitigate the spread of absent labels.  We report results on two datasets that illustrate two contrasting cases: in BACH, nearly all classes are present, whereas in CATCH only 4 of the 12 classes are present on average. On CATCH, the step schedule reduces the number of predicted classes while slightly increasing balanced accuracy and macro F1. On BACH, where absent labels are not a concern, performance is essentially unaffected. These behaviors are also visible qualitatively in Figure~\ref{fig:ablation_alpha}.

\vspace{2mm}

Performance when the class-presence term $\lambda$ is added is also reported. This term is complementary to $\alpha$ non-uniform, but is derived from the annotations and acts on the unary potential. This mechanism provides a substantial additional improvement, increasing balanced accuracy and macro F1 on both datasets, with $+12.3\%$ and $+3.3\%$ on CATCH, respectively, while reducing the number of predicted classes.

\subsubsection{Limitation of class-presence term}
\begin{figure}[!t]
    \centering
    \includegraphics[width=\linewidth]{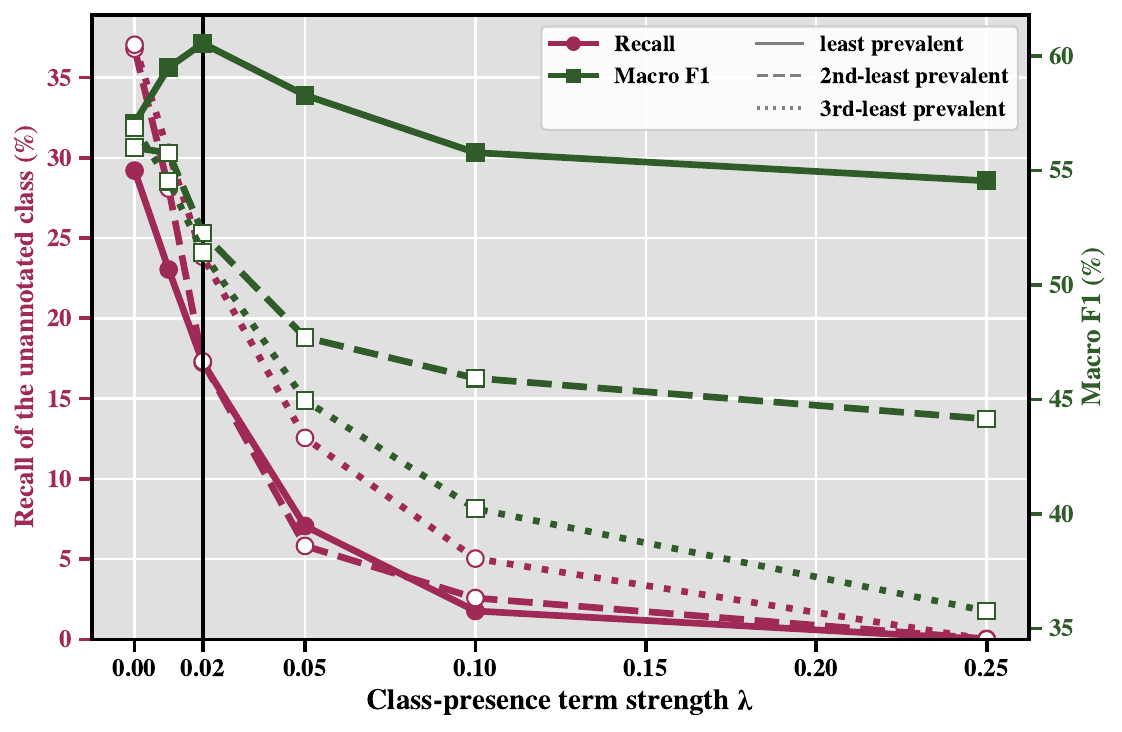}
    \caption{Effect of the class-presence term strength $\lambda$ on the macro F1 when a class in the slide is not annotated (least-, 2nd-least- and 3rd-least prevalent). It is important to notice that a moderate strength improves macro F1 without erasing an unannotated class.} 
    \label{fig:miss_class}
\end{figure}

To reflect a clinical setting in which a pathologist may overlook a small region, we evaluate the case where a class present in the slide is left unannotated. On CATCH, we omit the least, 2nd-least and 3rd-least prevalent class from the annotations and
measure how the class-presence term affects the recall of the unannotated class and the macro F1. As shown in Figure~\ref{fig:miss_class}, at the chosen strength ($\lambda=0.02$), the unannotated class is never erased and retains $17$--$24\%$
of its recall%, giving up $13.1$, $18.5$ and $12.6$ recall points for the least, 2nd-least and 3rd-least present class, respectively
, while macro F1 stays high. Suppression becomes severe only for $\lambda\geq0.1$, where the unannotated class collapses to $5\%$ recall or below, and vanishes entirely at $\lambda=0.25$.

\input{Tables/table_ablation_potentials}

\input{Tables/table_ablation_alpha}

\subsection{HITL interaction yields the best performance}
\begin{figure}[!h]
      \centering
      \includegraphics[width=\linewidth]{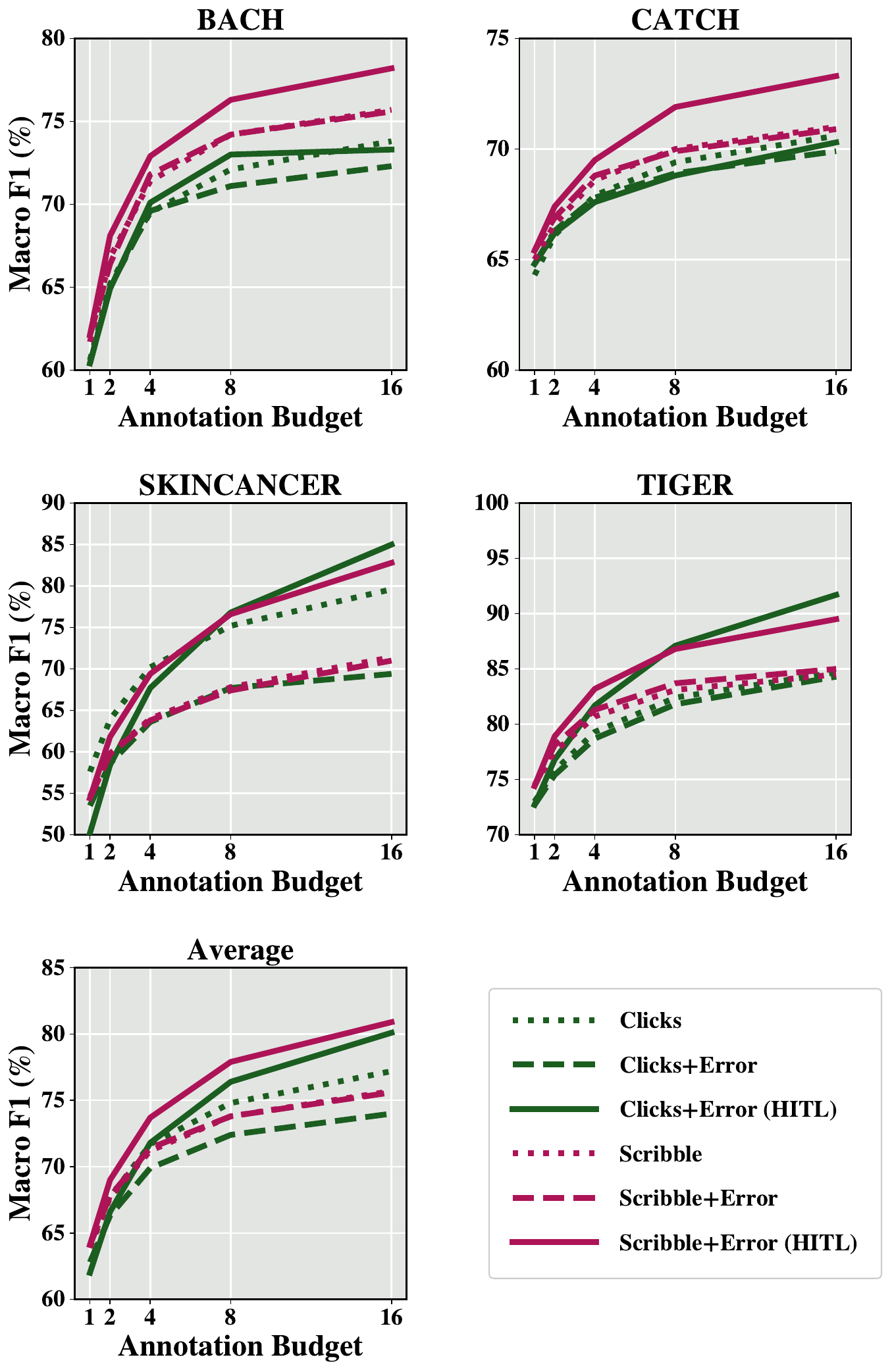}
      \caption{Macro F1 (\%) of the considered annotation protocols with varying annotation budgets $b \in \{1,2,4,8,16\}$ across four datasets and their average. Lines share a color per annotation type and a style per interaction type.}
      \label{fig:annotation_bacc_f1}
\end{figure}

The results of the different annotation protocols are summarized in Figure~\ref{fig:annotation_bacc_f1}. 
On most of the datasets, \emph{scribbles} match or outperform \emph{clicks}, most clearly in macro F1 at high budget: unlike a spatially compact click, a scribble spans a wider range of tissue structures and thus covers more diverse patches. For the same annotation budget, i.e.,\ the same number of pathologist actions, this provides richer supervision and improves propagation.

\vspace{2mm}

We then investigate whether correcting errors iteratively through a \emph{HITL} procedure is more effective than annotating the full budget at once. On every dataset, clicks-error \emph{HITL} increasingly outperforms \emph{clicks-error} as the budget grows, reaching $+6.1\%$ in macro F1 on average at 16 annotations per class. % (up to $+7.4\%$ on TIGER). 
\emph{HITL} continuously retargets the residual errors of the current, progressively improved prediction, allowing it to correct systematic confusions. 
This iterative benefit also extends to scribbles: scribble-error \emph{HITL} improves over single-step \emph{scribble-error} by $+5.3\%$ in macro F1 on average at 16 annotation actions. 

\vspace{2mm}

\textbf{Practical recommendation.} On average, iterative error correction (\emph{HITL}) yields the best performance, followed by single-step \emph{representative} annotation, with single-step \emph{error-driven} annotation ranking last. As for the choice of modality, \emph{scribbles} are the preferable default, matching or surpassing clicks on most datasets at a comparable budget.

%% file: Tables/table_bacc_f1.tex
\begin{table*}[h!]
\centering
 \caption{%
      Balanced accuracy (\%) and macro F1 (\%) for WSI patch classification with varying annotation budgets $b \in \{1,2,4,8,16\}$ \textit{clicks} (in the \textit{representative} interaction) across four datasets and their average.
      \textbf{ZS} is the CONCH zero-shot baseline.
      \textbf{Bold}: best per column; \underline{underline}: second best.%
  }
\vspace{-2mm}
\label{tab:bacc_f1}
\resizebox{\linewidth}{!}{%
\begin{tabular}{@{}>{\centering\arraybackslash}p{2em}clc c@{}c@{\hspace{1em}}ccccc c@{}c@{\hspace{1em}}ccccc@{}}
& & & & & & \multicolumn{5}{c}{\textbf{Balanced accuracy (\%)}} & & & \multicolumn{5}{c}{\textbf{Macro F1 (\%)}} \\
\cmidrule[\heavyrulewidth](lr){7-11} \cmidrule[\heavyrulewidth](lr){14-18}
\textbf{Dataset} & \textbf{Approach} & \textbf{Method} & & & \textbf{ZS} & \textit{1} & \textit{2} & \textit{4} & \textit{8} & \textit{16} & & \textbf{ZS} & \textit{1} & \textit{2} & \textit{4} & \textit{8} & \textit{16} \\
\midrule[\heavyrulewidth]
\multirow{7}{*}{\rotatebox[origin=c]{90}{\textbf{BACH}}} & \multirow{2}{*}{\textit{Inductive}} & Linear Probe & & \hspace{2.5em} & \multirow{2}{*}{41.8} & {52.5} & {53.9} & {57.5} & {64.3} & {69.0} & \hspace{2.5em} & \multirow{2}{*}{43.7} & {51.5} & {51.9} & {53.7} & {57.8} & {61.0} \\
 &  & Tip-Adapter-F & & \hspace{2.5em} &  & {49.2} & {53.9} & {60.2} & {67.0} & {73.8} & \hspace{2.5em} &  & {48.9} & {51.4} & {54.8} & {58.3} & {62.2} \\
\cmidrule(l){2-18}
 & \multirow{2}{*}{\textit{Probabilistic}} & Histo-TransCLIP & & \hspace{2.5em} &  \multirow{2}{*}{41.8} & {54.7} & {55.5} & {55.8} & {55.2} & {55.0} & \hspace{2.5em} &  \multirow{2}{*}{43.7} & {55.0} & {55.5} & {55.9} & {55.3} & {55.4} \\
 &  & PADDLE & & \hspace{2.5em} &  & {40.2} & {43.4} & {45.8} & {47.5} & {49.3} & \hspace{2.5em} &  & {34.9} & {38.5} & {41.1} & {42.7} & {44.3} \\
\cmidrule(l){2-18}
 & \multirow{3}{*}{\textit{Graph-based}} & ECALP & & \hspace{2.5em} &  \multirow{3}{*}{41.8} & \textbf{72.4} & \textbf{75.8} & \textbf{78.8} & \underline{80.5} & \underline{82.3} & \hspace{2.5em} &  \multirow{3}{*}{43.7} & \underline{59.4} & \underline{63.0} & \underline{65.7} & \underline{66.8} & \underline{69.9} \\
 &  & HistoCRF & & \hspace{2.5em} &  & {57.0} & {60.6} & {64.1} & {67.0} & {69.1} & \hspace{2.5em} &  & {44.7} & {46.6} & {49.0} & {50.4} & {51.6} \\
 &  & \cellcolor{hotpink}\textbf{SlideCRF} & \cellcolor{hotpink} & \cellcolor{hotpink}\hspace{2.5em} & \cellcolor{hotpink} & \cellcolor{hotpink}\underline{65.1} & \cellcolor{hotpink}\underline{71.1} & \cellcolor{hotpink}\underline{77.6} & \cellcolor{hotpink}\textbf{82.2} & \cellcolor{hotpink}\textbf{84.7} & \cellcolor{hotpink}\hspace{2.5em} & \cellcolor{hotpink} & \cellcolor{hotpink}\textbf{60.6} & \cellcolor{hotpink}\textbf{65.2} & \cellcolor{hotpink}\textbf{69.5} & \cellcolor{hotpink}\textbf{72.1} & \cellcolor{hotpink}\textbf{73.8} \\
\midrule[\heavyrulewidth]
\midrule
\multirow{7}{*}{\rotatebox[origin=c]{90}{\textbf{CATCH}}} & \multirow{2}{*}{\textit{Inductive}} & Linear Probe & & \hspace{2.5em} & \multirow{2}{*}{39.0} & {50.5} & {51.4} & {56.9} & {64.6} & {71.3} & \hspace{2.5em} & \multirow{2}{*}{36.9} & {45.8} & {46.2} & {50.4} & {56.0} & {61.5} \\
 &  & Tip-Adapter-F & & \hspace{2.5em} &  & {44.9} & {51.0} & {60.5} & {70.1} & {76.4} & \hspace{2.5em} &  & {43.0} & {49.2} & {58.7} & {66.8} & \underline{71.7} \\
\cmidrule(l){2-18}
 & \multirow{2}{*}{\textit{Probabilistic}} & Histo-TransCLIP & & \hspace{2.5em} &  \multirow{2}{*}{39.0} & {39.6} & {42.0} & {43.1} & {43.7} & {43.9} & \hspace{2.5em} &  \multirow{2}{*}{36.9} & {40.1} & {42.2} & {43.3} & {43.7} & {43.8} \\
 &  & PADDLE & & \hspace{2.5em} &  & {17.6} & {18.4} & {19.8} & {21.5} & {23.2} & \hspace{2.5em} &  & {22.2} & {23.5} & {25.2} & {27.2} & {29.0} \\
\cmidrule(l){2-18}
 & \multirow{3}{*}{\textit{Graph-based}} & ECALP & & \hspace{2.5em} &  \multirow{3}{*}{39.0} & \textbf{76.7} & \textbf{81.6} & \textbf{84.2} & \textbf{86.0} & \textbf{87.0} & \hspace{2.5em} &  \multirow{3}{*}{36.9} & \underline{63.5} & \textbf{67.0} & \textbf{69.7} & \textbf{71.9} & \textbf{74.0} \\
 &  & HistoCRF & & \hspace{2.5em} &  & {50.7} & {55.1} & {59.0} & {62.5} & {64.4} & \hspace{2.5em} &  & {46.5} & {49.4} & {52.3} & {54.9} & {56.6} \\
 &  & \cellcolor{hotpink}\textbf{SlideCRF} & \cellcolor{hotpink} & \cellcolor{hotpink}\hspace{2.5em} & \cellcolor{hotpink} & \cellcolor{hotpink}\underline{76.0} & \cellcolor{hotpink}\underline{78.3} & \cellcolor{hotpink}\underline{80.5} & \cellcolor{hotpink}\underline{82.6} & \cellcolor{hotpink}\underline{84.1} & \cellcolor{hotpink}\hspace{2.5em} & \cellcolor{hotpink} & \cellcolor{hotpink}\textbf{64.3} & \cellcolor{hotpink}\underline{66.1} & \cellcolor{hotpink}\underline{67.9} & \cellcolor{hotpink}\underline{69.4} & \cellcolor{hotpink}{70.6} \\
\midrule[\heavyrulewidth]
\midrule
\multirow{7}{*}{\rotatebox[origin=c]{90}{\textbf{SKINCANCER}}} & \multirow{2}{*}{\textit{Inductive}} & Linear Probe & & \hspace{2.5em} & \multirow{2}{*}{14.8} & {33.5} & {40.8} & {49.7} & {58.8} & {66.8} & \hspace{2.5em} & \multirow{2}{*}{9.0} & {12.8} & {13.2} & {13.9} & {15.4} & {17.0} \\
 &  & Tip-Adapter-F & & \hspace{2.5em} &  & {32.6} & {43.1} & {55.9} & {66.9} & {74.9} & \hspace{2.5em} &  & {13.7} & {19.6} & {27.2} & {33.6} & {37.5} \\
\cmidrule(l){2-18}
 & \multirow{2}{*}{\textit{Probabilistic}} & Histo-TransCLIP & & \hspace{2.5em} &  \multirow{2}{*}{14.8} & {25.6} & {25.9} & {26.1} & {26.1} & {25.9} & \hspace{2.5em} &  \multirow{2}{*}{9.0} & {19.2} & {19.1} & {19.1} & {19.0} & {18.9} \\
 &  & PADDLE & & \hspace{2.5em} &  & {41.4} & {48.3} & {55.5} & {61.8} & {67.4} & \hspace{2.5em} &  & {38.3} & {45.6} & {52.9} & {59.1} & {64.6} \\
\cmidrule(l){2-18}
 & \multirow{3}{*}{\textit{Graph-based}} & ECALP & & \hspace{2.5em} &  \multirow{3}{*}{14.8} & {54.4} & {61.1} & {66.0} & {68.5} & {69.9} & \hspace{2.5em} &  \multirow{3}{*}{9.0} & {47.5} & \underline{55.2} & \underline{61.2} & \underline{65.1} & {67.5} \\
 &  & HistoCRF & & \hspace{2.5em} &  & \underline{60.6} & \underline{67.7} & \underline{74.3} & \underline{79.3} & \underline{83.3} & \hspace{2.5em} &  & \underline{48.0} & {53.2} & {58.5} & {63.3} & \underline{67.6} \\
 &  & \cellcolor{hotpink}\textbf{SlideCRF} & \cellcolor{hotpink} & \cellcolor{hotpink}\hspace{2.5em} & \cellcolor{hotpink} & \cellcolor{hotpink}\textbf{68.8} & \cellcolor{hotpink}\textbf{75.4} & \cellcolor{hotpink}\textbf{81.4} & \cellcolor{hotpink}\textbf{85.4} & \cellcolor{hotpink}\textbf{88.4} & \cellcolor{hotpink}\hspace{2.5em} & \cellcolor{hotpink} & \cellcolor{hotpink}\textbf{57.6} & \cellcolor{hotpink}\textbf{63.9} & \cellcolor{hotpink}\textbf{70.2} & \cellcolor{hotpink}\textbf{75.2} & \cellcolor{hotpink}\textbf{79.6} \\
\midrule[\heavyrulewidth]
\midrule
\multirow{7}{*}{\rotatebox[origin=c]{90}{\textbf{TIGER}}} & \multirow{2}{*}{\textit{Inductive}} & Linear Probe & & \hspace{2.5em} & \multirow{2}{*}{74.7} & {77.5} & {77.8} & {78.6} & {80.0} & {81.4} & \hspace{2.5em} & \multirow{2}{*}{69.1} & {71.1} & {71.5} & {71.7} & {72.9} & {74.1} \\
 &  & Tip-Adapter-F & & \hspace{2.5em} &  & {76.5} & {77.7} & {79.7} & {81.8} & {84.0} & \hspace{2.5em} &  & {70.2} & {71.1} & {72.6} & {74.3} & {76.2} \\
\cmidrule(l){2-18}
 & \multirow{2}{*}{\textit{Probabilistic}} & Histo-TransCLIP & & \hspace{2.5em} &  \multirow{2}{*}{74.7} & \underline{80.5} & \underline{80.8} & {80.7} & {80.6} & {80.5} & \hspace{2.5em} &  \multirow{2}{*}{69.1} & \textbf{74.1} & {74.3} & {74.4} & {74.4} & {74.4} \\
 &  & PADDLE & & \hspace{2.5em} &  & {72.5} & {76.3} & {78.7} & {80.0} & {81.0} & \hspace{2.5em} &  & {64.2} & {68.3} & {71.4} & {72.9} & {74.2} \\
\cmidrule(l){2-18}
 & \multirow{3}{*}{\textit{Graph-based}} & ECALP & & \hspace{2.5em} &  \multirow{3}{*}{74.7} & \textbf{81.6} & \textbf{83.8} & \textbf{85.4} & \textbf{86.4} & \textbf{87.4} & \hspace{2.5em} &  \multirow{3}{*}{69.1} & \underline{73.5} & \underline{75.7} & \underline{77.7} & \underline{79.4} & \underline{81.0} \\
 &  & HistoCRF & & \hspace{2.5em} &  & {76.9} & {77.5} & {78.3} & {78.7} & {79.2} & \hspace{2.5em} &  & {67.0} & {67.5} & {68.0} & {68.0} & {67.9} \\
 &  & \cellcolor{hotpink}\textbf{SlideCRF} & \cellcolor{hotpink} & \cellcolor{hotpink}\hspace{2.5em} & \cellcolor{hotpink} & \cellcolor{hotpink}{76.4} & \cellcolor{hotpink}{78.9} & \cellcolor{hotpink}\underline{82.3} & \cellcolor{hotpink}\underline{85.0} & \cellcolor{hotpink}\underline{87.1} & \cellcolor{hotpink}\hspace{2.5em} & \cellcolor{hotpink} & \cellcolor{hotpink}{73.0} & \cellcolor{hotpink}\textbf{75.8} & \cellcolor{hotpink}\textbf{79.3} & \cellcolor{hotpink}\textbf{82.4} & \cellcolor{hotpink}\textbf{84.7} \\
\specialrule{.3em}{.3em} {.3em}
\multirow{7}{*}{\rotatebox[origin=c]{90}{\textbf{AVERAGE}}} & \multirow{2}{*}{\textit{Inductive}} & Linear Probe & & \hspace{2.5em} & \multirow{2}{*}{42.6} & {53.5} & {56.0} & {60.6} & {66.9} & {72.1} & \hspace{2.5em} & \multirow{2}{*}{39.7} & {45.3} & {45.7} & {47.4} & {50.5} & {53.4} \\
 &  & Tip-Adapter-F & & \hspace{2.5em} &  & {50.8} & {56.4} & {64.1} & {71.4} & {77.3} & \hspace{2.5em} &  & {43.9} & {47.8} & {53.3} & {58.3} & {61.9} \\
\cmidrule(l){2-18}
 & \multirow{2}{*}{\textit{Probabilistic}} & Histo-TransCLIP & & \hspace{2.5em} &  \multirow{2}{*}{42.6} & {50.1} & {51.1} & {51.5} & {51.4} & {51.3} & \hspace{2.5em} &  \multirow{2}{*}{39.7} & {47.1} & {47.8} & {48.2} & {48.1} & {48.1} \\
 &  & PADDLE & & \hspace{2.5em} &  & {42.9} & {46.6} & {49.9} & {52.7} & {55.2} & \hspace{2.5em} &  & {39.9} & {44.0} & {47.7} & {50.5} & {53.1} \\
\cmidrule(l){2-18}
 & \multirow{3}{*}{\textit{Graph-based}} & ECALP & & \hspace{2.5em} &  \multirow{3}{*}{42.6} & \underline{71.3} & \underline{75.6} & \underline{78.6} & \underline{80.4} & \underline{81.7} & \hspace{2.5em} &  \multirow{3}{*}{39.7} & \underline{61.0} & \underline{65.2} & \underline{68.6} & \underline{70.8} & \underline{73.1} \\
 &  & HistoCRF & & \hspace{2.5em} &  & {61.3} & {65.2} & {68.9} & {71.9} & {74.0} & \hspace{2.5em} &  & {51.5} & {54.1} & {56.9} & {59.1} & {60.9} \\
 &  & \cellcolor{hotpink}\textbf{SlideCRF} & \cellcolor{hotpink} & \cellcolor{hotpink}\hspace{2.5em} & \cellcolor{hotpink} & \cellcolor{hotpink}\textbf{71.6} & \cellcolor{hotpink}\textbf{75.9} & \cellcolor{hotpink}\textbf{80.4} & \cellcolor{hotpink}\textbf{83.8} & \cellcolor{hotpink}\textbf{86.1} & \cellcolor{hotpink}\hspace{2.5em} & \cellcolor{hotpink} & \cellcolor{hotpink}\textbf{63.9} & \cellcolor{hotpink}\textbf{67.7} & \cellcolor{hotpink}\textbf{71.7} & \cellcolor{hotpink}\textbf{74.8} & \cellcolor{hotpink}\textbf{77.2} \\
\bottomrule
\end{tabular}}
\end{table*}

%% file: Tables/runtime.tex
\begin{table}[!t]
\centering
\footnotesize

\caption{Total runtime (in seconds) on an NVIDIA A10 GPU, as a function of the number of patches.}
\vspace{-1mm}
\label{tab:time_transposed}
\resizebox{\linewidth}{!}{%
\begin{tabular}{llccc}
\toprule
\textbf{Approach} & \textbf{Method} & \textbf{$\sim 10^{3}$ } & \textbf{$\sim 10^{4}$ } & \textbf{$\sim 10^{5}$ } \\
\midrule
\multirow{2}{*}{\textit{Inductive}}
  & Linear Probe   & 6.3 & 12.6 & 32.6 \\
  & Tip-Adapter-F  & 0.1 & 5.1  & 25.8 \\
\midrule
\multirow{2}{*}{\textit{Probabilistic}}
  & Histo-TransCLIP      & 0.1 & 1.0 & 4.4 \\
  & PADDLE         & 0.8 & 11.4 & 40.5 \\
\midrule
\multirow{2}{*}{\textit{Graph-based}}
  & ECALP          & 8.0 & 142.6 & 490.9 \\
  & \cellcolor{hotpink}\textbf{SlideCRF} & \cellcolor{hotpink}3.2 & \cellcolor{hotpink}12.6 & \cellcolor{hotpink}37.8 \\
\bottomrule
\end{tabular}%
}

\end{table}

%% file: Tables/table_ablation_potentials.tex
\begin{table}[!h]
\centering
\caption{Contribution of each term to the pairwise potential: the spatial and biological terms are added cumulatively on top of the annotation and diversity terms. %Incremental contribution of the spatial and biological terms to the pairwise potential. 
Balanced accuracy (\%) and macro F1 (\%) are averaged over the four datasets.}
\label{tab:ablation_potentials}
\resizebox{\linewidth}{!}{%
\begin{tabular}{lcc}
\toprule
\textbf{Pairwise terms} & \textbf{Balanced accuracy (\%)} & \textbf{Macro F1 (\%)} \\
\midrule
Annotation + Diversity & 74.3 & 58.2 \\
\hspace{3mm} + Spatial & 80.2 & 66.3 \\
\rowcolor{hotpink} \hspace{3mm} + Biological (SlideCRF) & 80.4 & 71.7 \\
\bottomrule
\end{tabular}%
}
\end{table}

%% file: Tables/table_ablation_alpha.tex
\begin{table}[!t]
\centering
\caption{Contribution of the weighting factor $\alpha$ and the class-presence term $\lambda$. $n_{pred}/n_{gt}$ = avg.\ predicted vs.\ present classes per slide.}
\label{tab:ablation_alpha}
\resizebox{\linewidth}{!}{%
\begin{tabular}{llcccc} % Remade to a clean 6-column format; see note below if keeping hotpink
\toprule
\textbf{Dataset} & \textbf{Metric} & $\boldsymbol{\alpha=0}$ & $\boldsymbol{\alpha}$ \textbf{uniform} & $\boldsymbol{\alpha}$ \textbf{non-uniform} & \cellcolor{hotpink}$\boldsymbol{+ \lambda \neq 0}$ \\
\midrule
 & Bal. Acc & 42.3 & 68.2 & 68.2 & \cellcolor{hotpink}80.5 \\
\textbf{CATCH} & Macro F1 & 38.1 & 63.2 & 64.6 & \cellcolor{hotpink}67.9 \\
               & $n_{pred}/n_{gt}$ & 4.1/3.7 & 12.9/3.7 & 10.1/3.7 & \cellcolor{hotpink}7.6/3.7 \\
\midrule
               & Bal. Acc & 52.2 & 74.3 & 72.8 & \cellcolor{hotpink}77.6 \\
\textbf{BACH}  & Macro F1 & 49.4 & 66.4 & 66.9 & \cellcolor{hotpink}69.5 \\
               & $n_{pred}/n_{gt}$ & 3.5/3.4 & 4.0/3.4 & 3.9/3.4 & \cellcolor{hotpink}3.4/3.4 \\
\bottomrule
\end{tabular}%
} 
\end{table}

%% file: Sections/5-conclusion.tex
We introduce SlideCRF, a CRF tailored to a realistic clinical setting of WSI analysis, where a few expert annotations and the relations between patches are exploited to refine patch-level predictions. To account for the spatial organization and the severe class imbalance specific to WSIs, our formulation combines terms integrating spatial and biological information with a term that penalizes the prediction of classes that are never annotated. Across four diverse datasets, SlideCRF outperforms strong baselines in both balanced accuracy and macro F1, highlighting its potential for clinically realistic WSI refinement. Our annotation study further shows that iterative error correction yields the largest gains, confirming the value of keeping the pathologist in the loop.\\

\textbf{Future Work.} Several directions remain open. A natural next step is to validate the proposed annotation protocols directly with pathologists, in order to confirm their clinical realism. Efficiency and accuracy could also be improved by exploiting patches at multiple resolutions, following the pyramidal structure of WSIs. Beyond histology, our framework could be extended to 3D medical images, where volumes exhibit analogous spatial connectivity. 
More broadly, quantifying how each protocol affects refinement provides a common ground for comparing future interactive annotation methods, which is essential before deploying them in routine clinical practice, where pathologist annotation time is scarce and directly constrains diagnostic throughput.